\documentclass[letterpaper, 10 pt, conference]{ieeeconf}  

\IEEEoverridecommandlockouts                              

\usepackage{cite}
\usepackage{amsmath,amssymb,amsfonts}
\usepackage{algorithmic}
\usepackage{graphicx}
\usepackage{textcomp}
\usepackage{xcolor}
\usepackage{svg}
\usepackage{svg-extract}
\usepackage{dsfont}
\usepackage{adjustbox}

\usepackage{xspace}
\usepackage{tikz}
\usepackage{hyperref}

\newcommand\copyrighttext{%
  \footnotesize \textcopyright 2025 IEEE. Personal use of this material is permitted.
  Permission from IEEE must be obtained for all other uses, in any current or future
  media, including reprinting/republishing this material for advertising or promotional
  purposes, creating new collective works, for resale or redistribution to servers or
  lists, or reuse of any copyrighted component of this work in other works.}
\newcommand\copyrightnotice{%
\begin{tikzpicture}[remember picture,overlay]
\node[anchor=south,yshift=10pt] at (current page.south) 
  {\fbox{\parbox{\dimexpr\textwidth-\fboxsep-\fboxrule\relax}{\copyrighttext}}};
\end{tikzpicture}%
}

\providecommand{\Comma}{\text{~,\xspace}}

\title{\LARGE \bf
Near Time-Optimal Hybrid Motion Planning for Timber Cranes
}

\author{Marc-Philip Ecker$^{1,2}$, Bernhard Bischof$^{2}$, Minh Nhat Vu$^{1,2}$, Christoph Fröhlich$^{2}$,\\ Tobias Glück$^{2}$ and Wolfgang Kemmetmüller$^{1}$
\thanks{$^{1}$Marc-Philip Ecker, Minh Nhat Vu and Wolfgang Kemmetmüller are with the
Automation \& Control Institute (ACIN), TU Wien, 1040 Vienna, Austria
        {\tt\small \{ecker,vu,kemmetmueller\}@acin.tuwien.ac.at}}%
\thanks{$^{2}$Marc-Philip Ecker, Bernhard Bischof, Minh Nhat Vu, Christoph Fröhlich and Tobias Glück are with the Center for Vision, Automation \& Control,
AIT Austrian Institute of Technology GmbH, 1210 Vienna, Austria
        {\tt\small \{marc-philip.ecker,bernhard.bischof,minh.vu,\newline christoph.froehlich,tobias.glueck\}@ait.ac.at}}%
}

\begin{document}
\maketitle
\copyrightnotice
\thispagestyle{empty}
\pagestyle{empty}
\begin{abstract}

Efficient, collision-free motion planning is essential for automating large-scale manipulators like timber cranes. They come with unique challenges such as hydraulic actuation constraints and passive joints—factors that are seldom addressed by current motion planning methods. This paper introduces a novel approach for time-optimal, collision-free hybrid motion planning for a hydraulically actuated timber crane with passive joints. We enhance the via-point-based stochastic trajectory optimization (VP-STO) algorithm to include pump flow rate constraints and develop a novel collision cost formulation to improve robustness. The effectiveness of the enhanced VP-STO as an optimal single-query global planner is validated by comparison with an informed RRT* algorithm using a time-optimal path parameterization (TOPP). The overall hybrid motion planning is formed by combination with a gradient-based local planner that is designed to follow the global planner's reference and to systematically consider the passive joint dynamics for both collision avoidance and sway damping.  
\end{abstract}

\section{Introduction}
Cranes are a pivotal automation component in modern forestry, enhancing the efficiency and safety of timber handling and transportation. The process begins with harvesting, where trees are cut, delimbed, and processed into logs using harvesters. ers with cranes collect and transport the logs to a roadside landing. Cranes then load the logs onto timber trucks for transport to processing facilities, where cranes at the mill unload them for further processing. Due to an aging workforce and high training costs, there is an increasing shortage of skilled operators, which highlights the urgent need for autonomous or semi-autonomous systems, see~\cite{agostini:2003,starr:2005,morales:2014,kalmari:2014}. 
Timber cranes are underactuated large-scale manipulators that operate in unstructured outdoor environments, making collision-free motion planning essential. These cranes typically use hydraulic actuators to meet high-force requirements. The hydraulic power supply is provided by a pump with a limited maximum pump flow rate. This induces coupled limits on the maximum speed of the individual joint of the manipulator. Therefore, effective motion planning must consider these hydraulic limitations to ensure fast, collision-free operations. Various approaches to develop (semi-)autonomous large-scale manipulators have recently been proposed, cf. \cite{kalmari:2017,song:2020,dhakate:2022,jebellat:2023}. However, none of these works specifically address the challenge of time-optimal and collision-free point-to-point motion planning while considering the limitations of hydraulic actuators.


\begin{figure}[t]
\centering
\includegraphics[trim=10.0cm 10cm 10.0cm 0cm,clip,scale=0.25]{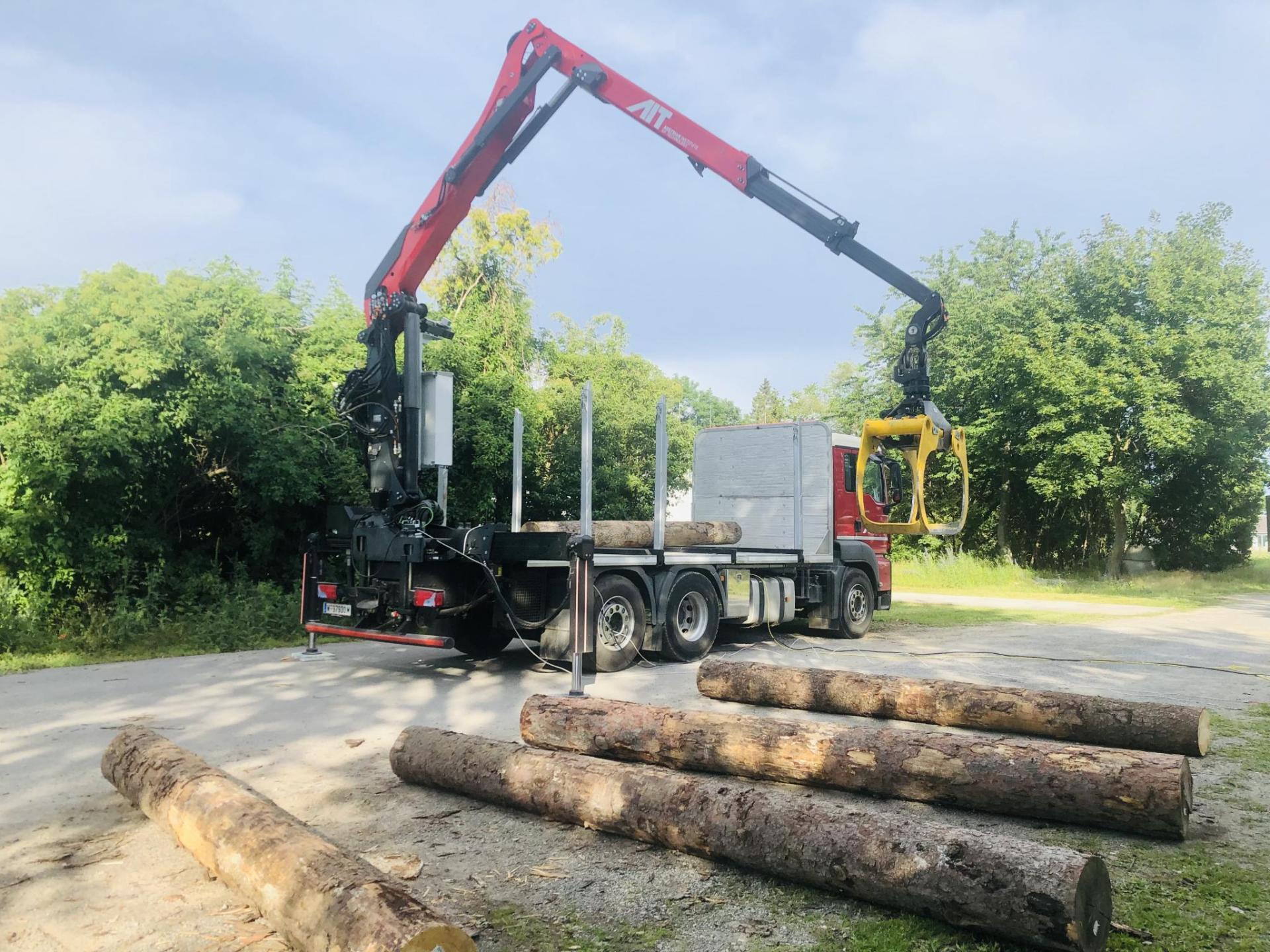}
\caption{The actual timber crane considered in this work.}
\label{fig:KinematicChain}
\end{figure}

\subsection{Related Work}
Motion planning involves generating a feasible trajectory from start to end while considering kinematic and dynamic constraints. Fully-actuated manipulators enable direct control of each joint, which simplifies motion planning. In contrast, underactuated manipulators, with fewer actuators than degrees of freedom, present more complex planning challenges due to passive dynamics and limited control authority. For hydraulically driven cranes with passive joints, MPC-based tracking control is the most commonly used method \cite{morales:2014,kalmari:2014,kalmari:2017,jebellat:2023}. However, these studies assume the availability of a global, collision-free path, thereby lacking the capability for autonomous operation in challenging environments.

In contrast, Ayoub et al. \cite{ayoub:2024} employ an RRT* planner to compute joint trajectories that avoid both self-collisions and collisions with the truck. However, the focus of \cite{ayoub:2024} is on the overall autonomous log loading scenario rather than on collision-free motion planning specifically for the crane. Consequently, their approach does not account for additional environmental obstacles, time-optimal planning with pump flow constraints, or the dynamics of passive joints.

Morales et al. \cite{morales:2014} provide a trajectory planning and motion control framework for the log loading task of a timber crane. The approach relies on pre-recorded paths from human operations to avoid collisions with the truck, but additional obstacles in the environment are not considered.
In \cite{kalmari:2014, kalmari:2017}, a nonlinear MPC for an underactuated timber crane is proposed with application to tip path tracking, utilizing predefined paths. 
Anti-sway motion generation for underactuated manipulators is considered in \cite{agostini:2003, starr:2005, jebellat:2023}.

Based on this overview of work for the specific application in timber cranes, we want to give a brief, more general overview of state-of-the-art motion planning techniques.
\subsubsection{Sampling-based planners}
Sampling-based planners (SBPs) \cite{OrtheyChamzasKavraki2024}, such as Probabilistic Roadmaps (PRM) \cite{kavraki:1996}, Rapidly-exploring Random Trees (RRT) \cite{lavalle:1999} and their extensions \cite{Karaman2011,gammel:2014,Gammell2020},
are known for their probabilistic completeness properties. However, incorporating differential constraints significantly increases the dimensionality of the search space, leading to an exponential increase in the number of required samples \cite{Elbanhawi2014}. 
\subsubsection{Trajectory optimization}
Trajectory optimization (TO) involves generating an optimal trajectory by solving an optimal control problem (OCP) that satisfies kinodynamic and task-specific constraints. Gradient-based optimization approaches, such as CHOMP \cite{ratliff:2009}, TrajOpt \cite{schulman:2014}, and ALTRO \cite{howell:2019}, are specific methods within the broader category of TO. These approaches use gradient information to iteratively improve the trajectory.
Unlike sampling-based planning methods, TO efficiently incorporates dynamic feasibility, kinematic reachability, and collision constraints simultaneously. However, highly nonconvex constraints, such as those encountered in collision avoidance, can cause gradient-based optimization approaches to become trapped in local minima \cite{dai:2018}.

\subsubsection{Hybrid methods}
Hybrid methods \cite{park:2015,li:2016,dai:2018,leu:2021,leu:2022,jelavic:2023,kamat:2022} use the solutions of sampling-based planners to generate an initial feasible path and then apply trajectory optimization to refine this path into a time-optimal or smooth trajectory. The methods proposed in \cite{li:2016,dai:2018,leu:2021,leu:2022,kamat:2022} use TO to improve the smoothness as well as the path length with application to fully actuated manipulators.
Contrary, the works \cite{volz:2019,jelavic:2023} use SBPs as global planners to get a reference path that is used as guidance for the TO-based local planners, which incorporates dynamic constraints neglected in the global planner. Our approach follows the latter category of hybrid motion planning.

\subsubsection{Stochastic Trajectory Optimization}
Another promising solution to avoid getting stuck in local minima due to collision constraints are stochastic trajectory optimization (STO) methods, such as Model Predictive Path Integral control \cite{kappen:2005, williams:2017}.
However, as noted by \cite{janakowski:2023}, these methods are typically limited to short-horizon problems. Via-point Stochastic Trajectory Optimization (VP-STO) \cite{janakowski:2023} addresses this limitation by optimizing over a continuous trajectory space defined by via-points. Moreover, hard constraints on fixed start and end configurations, as well as velocity and acceleration limits, are implicitly handled by the trajectory representation. Therefore, VP-STO is well-suited for computing full-horizon trajectories.

\subsection{Contribution}

This paper introduces a hybrid motion planning framework for hydraulically actuated timber cranes with two passive joints. 
To achieve this, we:
\begin{itemize}
\item Extend VP-STO \cite{janakowski:2023} to handle the Pump Flow Rate Constraint (PFRC), creating PFRC-VP-STO, which enables the planning of near time-optimal and collision-free trajectories for hydraulic manipulators.
\item Demonstrate the potential of PFRC-VP-STO as an optimal single-query global planner for the timber crane, and propose a novel collision cost formulation to enhance robustness.
\item Propose a hybrid motion planning method, combining the PFRC-VP-STO with an iLQR-based local planner, to achieve near time-optimal motions by systematically considering the passive joint dynamics for both collision avoidance and sway damping.
\end{itemize}
To the best of our knowledge, this is the first work addressing hybrid motion planning specifically for timber cranes. We bridge the gap left by previous studies \cite{morales:2014,kalmari:2014,kalmari:2017,jebellat:2023} by proposing a strategy to compute a global collision-free and time-optimal reference trajectory in complex environments, which is tracked by a subordinate local planner.



\subsection{Paper structure}
The remainder of this paper is structured as follows:
Section~\ref{sec:Model} introduces the mathematical model of the timber crane considered in this work. Section~\ref{sec:Approach} details the core contributions of the proposed hybrid motion planning framework. The numerical results are presented in Section~\ref{sec:Results}. Finally, Section~\ref{sec:conclusion} concludes the paper and provides an outlook on future work.


\section{Mathematical modeling}\label{sec:Model}
We consider a configuration space ($\mathcal{C}$-space) $\mathcal{C}=\mathcal{C}_A\times\mathcal{C}_P\subseteq\mathds{R}^{n}$ that can be partitioned into an actuated $\mathcal{C}$-space $\mathcal{C}_A\in\mathds{R}^{n_A}$ and a passive $\mathcal{C}$-space $\mathcal{C}_P\in\mathds{R}^{n_P}$. Therefore, a configuration $\mathbf{q}\in\mathcal{C}$ is represented as
\begin{align}\label{eq:PartitionGeneralizedCoordinates}
    \mathbf{q}^{\mathrm{T}}=\begin{bmatrix}
        \mathbf{q}_A^{\mathrm{T}}&
        \mathbf{q}_P^{\mathrm{T}}
    \end{bmatrix}\in\mathcal{C}=\mathcal{C}_A\times\mathcal{C}_P \ .
\end{align}
\begin{figure}[t]
\centering
\includegraphics[trim=0cm 0cm 0cm 0cm,clip,scale=0.44]{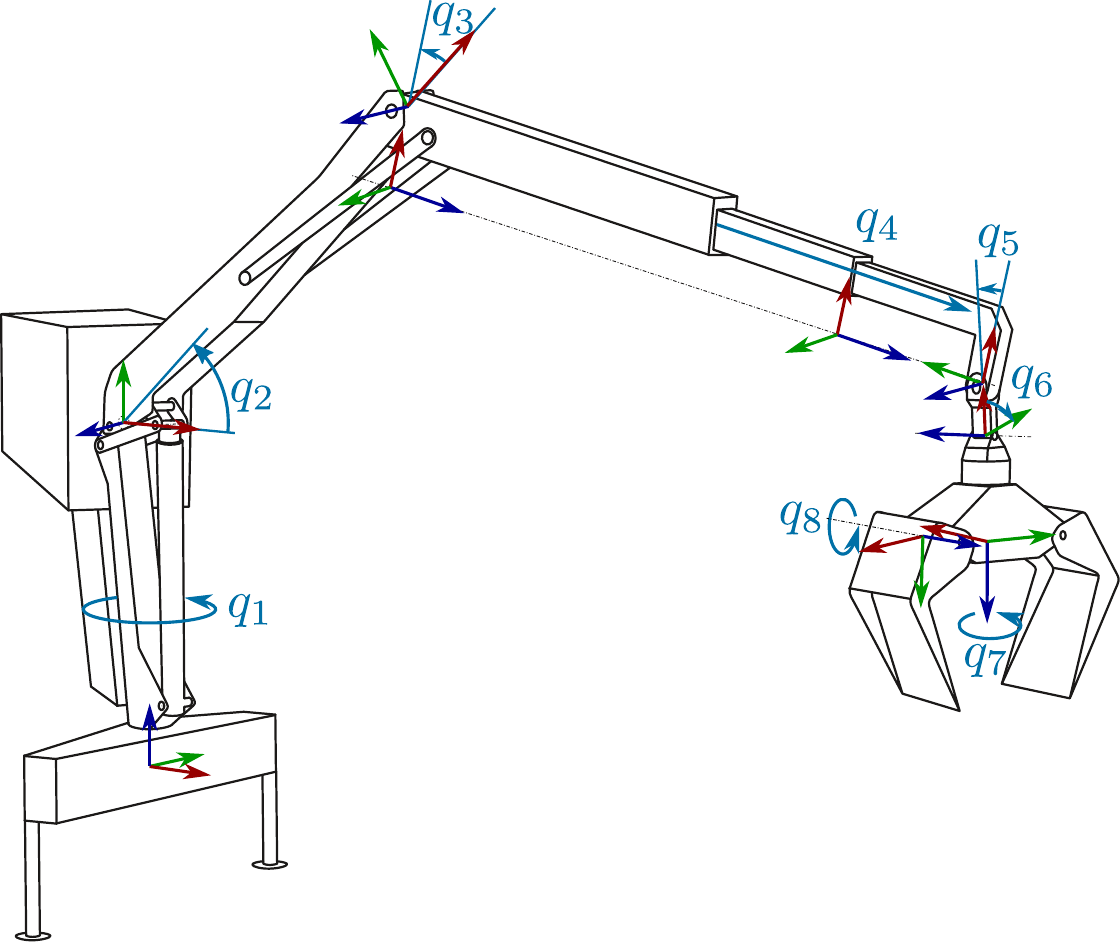}
\caption{Illustration of the kinematic chain of the timber crane, where $q_1,q_2,q_3,q_4,q_7$ and $q_8$ are actuated and $q_5$ and $q_6$ are passive joints.}
\label{fig:KinematicChain}
\end{figure}
Figure~\ref{fig:KinematicChain} illustrates the kinematic chain of the timber crane. We reasonably assume that the opening angle of the jaw, $q_8$, remains constant during point-to-point planning. This assumption reduces the jaw to a single rigid body, where the mass parameters of the wood log are incorporated into the jaw. Thus, we have $\mathbf{q}^{\mathrm{T}}_A=[q_1,q_2,q_3,q_4,q_7]$ and $\mathbf{q}_P^{\mathrm{T}}=[q_5,q_6]$. Notably, the displacements of the inner and outer telescope arms are coupled, hence represented by a single generalized coordinate $q_4$.

\subsection{Dynamic model}
The rigid body dynamic equations of the underactuated manipulator are given by
\begin{align}\label{eq:RigidBodyDynamics}
\begin{bmatrix}
        \mathbf{D}_{A}(\mathbf{q}) \!& \!\mathbf{D}^{\mathrm{T}}_{M}(\mathbf{q})\\
        \mathbf{D}_{M}(\mathbf{q})\! & \!\mathbf{D}_{P}(\mathbf{q})
    \end{bmatrix}
\ddot{\mathbf{q}}\! + \!\begin{bmatrix}
        \mathbf{C}_A(\mathbf{q},\dot{\mathbf{q}})\\
        \mathbf{C}_P(\mathbf{q},\dot{\mathbf{q}})
    \end{bmatrix}\dot{\mathbf{q}}\! + \!\begin{bmatrix}
        \mathbf{g}_A(\mathbf{q})\\
        \mathbf{g}_P(\mathbf{q})
    \end{bmatrix}\! = \!\begin{bmatrix}
        \boldsymbol{\tau}_A\\
        \mathbf{0}
    \end{bmatrix} \ ,
\end{align}
where we partitioned the mass matrix into $\mathbf{D}_{A}(\mathbf{q})\in\mathds{R}^{n_A\times n_A}$, $\mathbf{D}_{M}(\mathbf{q})\in\mathds{R}^{n_P\times n_A}$, $\mathbf{D}_{P}(\mathbf{q})\in\mathds{R}^{n_P\times n_P}$, the Coriolis matrix into $\mathbf{C}_A(\mathbf{q},\dot{\mathbf{q}})\in\mathds{R}^{n_A\times n}$, $\mathbf{C}_P\in\mathds{R}^{n_P\times n}$, and the potential forces into $\mathbf{g}_A(\mathbf{q})\in\mathbb{R}^{n_A}$ and $\mathbf{g}_P(\mathbf{q})\in\mathbb{R}^{n_P}$. The external forces are $\boldsymbol{\tau}_A\in\mathbb{R}^{n_A}$.

In the studied application, the hydraulic cylinders are controlled by decentralized subordinatecontrollers. Therefore, we can assume that the accelerations $\mathbf{u}=\ddot{\mathbf{q}}_A$ of the actuated subsystem are available as the (virtual) control input for the motion planning tasks. This results in the simplified model\begin{align}
    \ddot{\mathbf{q}}_P&=-\mathbf{D}_{P}^{-1}(\mathbf{q})(\mathbf{D}_{M}(\mathbf{q})\ddot{\mathbf{q}}_A+\mathbf{C}_P(\mathbf{q},\dot{\mathbf{q}})\dot{\mathbf{q}}+\mathbf{g}_P(\mathbf{q})) \label{eq:ddqu_DynamicsSimplified}
\end{align}
The equilibrium point of the passive joints $\bar{\mathbf{q}}_P\in\mathcal{C}_P$ for a given actuated configuration $\mathbf{q}_A\in\mathcal{C}_A$ can be computed by solving $\mathbf{g}_P(\mathbf{q}_A,\bar{\mathbf{q}}_P)=\mathbf{0}$.
The solution is denoted as $\bar{\mathbf{q}}_P=\mathbf{g}_P^{-1}(\mathbf{q}_A)$.

\subsection{Pump flow rate}\label{sec:Total pump flow rate}
The displacements of the hydraulic cylinders denoted by $\mathbf{d}^{\mathrm{T}}=[{d}_1,\dots,{d}_{n_A}]\in\mathbb{R}^{n_A}$, are described as a function of $\mathbf{q}_A$ by $\mathbf{d}=\mathbf{h}_C(\mathbf{q}_A)$. This directly results in the differential kinematics 
\begin{align}\label{eq:HydraulicVelocities}
    \dot{\mathbf{d}}&=\mathbf{J}_C(\mathbf{q}_A)\dot{\mathbf{q}}_A=\frac{\partial\mathbf{h}_C(\mathbf{q}_A)}{\partial\mathbf{q}_A}\dot{\mathbf{q}}_A \ ,
\end{align}
where $\dot{\mathbf{d}}\in\mathbb{R}^{n_A}$ are the velocities of the hydraulic actuators and $\dot{\mathbf{q}}_A$ are the joint velocities. The cylinders are supplied by a single pump, resulting in a pump flow rate (PFR) for the hydraulic system given by
\begin{align}\label{eq:TotalPumpFlow}
    Q(t) &= \sum_{l=1}^{n_A}A_l\big(\mathrm{sign}(\dot{d}_l)\big)|\dot{d}_l| \ ,
\end{align}
where $A_l(\cdot)$, $l=1,\dots,n_A$ are the direction-dependent effective areas of the cylinders. The PFR (\ref{eq:TotalPumpFlow}) that can be supplied by the hydraulic actuation system is limited by $Q_{\mathrm{max}}$. This yields the pump flow rate constraint (PFRC) $Q(t)\leq Q_{\mathrm{max}}$.

\subsection{Collision bodies}\label{sec:CollisionBodies}
To incorporate collision constraints, a simplified geometrical model of the truck and crane is used.
The links $\mathcal{L}_{i}(\mathbf{q})$, $i=2,3$ connecting joints $q_2$, $q_3$ and $q_3$, $q_5$ respectively, as well as the wood log $\mathcal{L}_5(\mathbf{q})$ are modeled as capsules. Notably, the length of $\mathcal{L}_3(\mathbf{q})$ depends on $q_4$. The slewing column $\mathcal{L}_1$ and the jaw $\mathcal{L}_{4}(\mathbf{q})$ are represented by oriented bounding boxes (OBB). The truck is modeled by the OBBs $\mathcal{O}_i$, $i=1,\dots,N_{\mathrm{truck}}$. Additional obstacles in the environment are added as convex objects. Collisions with the environment are incorporated by adding the constraints
\begin{align}
    \mathrm{sd}\big(\mathcal{L}_i(\mathbf{q}),\mathcal{O}_j\big)> 0\Comma
\end{align}
$i\in\{2,3,4,5\}$ and $j\in\{1,\dots,N_{\mathrm{obs}}\}$, where $\mathrm{sd}$ is the signed distance function and $N_{\mathrm{obs}}=N_{\mathrm{truck}}+N_{\mathrm{env}}$ includes the number of obstacles representing the truck $N_{\mathrm{truck}}$ and optional environment obstacles $N_{\mathrm{env}}$. Additionally, collisions between the log and the crane are incorporated using
\begin{align}
    \mathrm{sd}\big(\mathcal{L}_5(\mathbf{q}),\mathcal{L}_i(\mathbf{q})\big)> 0\Comma
\end{align}
$i\in\{1,2,3\}$. For shorter notation, we introduce the set $\mathcal{I}_c$ of all collision-pairs $(i,j)\in\mathcal{I}_c$ and write the signed distance short as $\mathrm{sd}_{ij}(\mathbf{q})$.

\section{Hybrid motion planning framework}\label{sec:Approach}
Our hybrid motion planning framework initially employs stochastic trajectory optimization (VP-STO with PFRC) to generate a global guiding trajectory neglecting the dynamics of the passive joints. For collision checking, the equilibrium point $\bar{\mathbf{q}}_P$ is used. Subsequently, we use a local gradient-based planner that takes care of the motions of the passive joints due to (\ref{eq:ddqu_DynamicsSimplified}). 

\subsection{Global Planner}
In this work, we utilize VP-STO as a global planner that computes a time-optimal guidance trajectory for the local planner. We just give a brief overview of the algorithm and explain the proposed extensions. For a detailed discussion of the basics of VP-STO the reader is referred to \cite{janakowski:2023}.

\subsubsection{VP-STO}
A trajectory is represented by $N_{\mathrm{via}}$ via-points $\boldsymbol{\xi}^{\mathrm{T}}=[\mathbf{q}_{A,\mathrm{via},1}^{\mathrm{T}},\dots,\mathbf{q}_{A,\mathrm{via},N_{\mathrm{via}}}^{\mathrm{T}}]$. A path $\hat{\mathbf{q}}_A(s)$ is computed using cubic spline interpolation and the trajectory is obtained using a linear time parameterization of the path variable $s(t)=\frac{t}{T}$ with final time $T>0$. To determinate the final time $T$ and to evaluate the cost function, $N_{\mathrm{eval}}$ equidistant points $\hat{\mathbf{q}}_{A,k}$ are sampled along the path $\hat{\mathbf{q}}_A(s)$. With
\begin{align}\label{eq:TPVelAcc}
    \dot{\mathbf{q}}_A(t)&=\hat{\mathbf{q}}_A^\prime(s)\dot{s} \Comma & \ddot{\mathbf{q}}_A(t)&=\hat{\mathbf{q}}_A^{\prime\prime}(s)\dot{s}^2 + \hat{\mathbf{q}}_A^{\prime}(s)\ddot{s}
\end{align}
and $\dot{s}=\frac{1}{T}$, $\ddot{s}=0$, the velocity and acceleration limits yield inequality constraints of the form $T\geq T_i(\boldsymbol{\xi})$ for all evaluation points. The final time is chosen as $T(\boldsymbol{\xi})=\max_iT_i(\boldsymbol{\xi})$, which corresponds to the minimal trajectory duration that satisfies all constraints. Hence, the final time $T(\boldsymbol{\xi})$ as well as the evaluation points along the trajectory are all uniquely defined by the via-points $\boldsymbol{\xi}$, which serve as optimization variables. The CMA-ES algorithm \cite{hansen:2023}, a stochastic black-box optimization algorithm, is utilized to solve
\begin{align}
    \min_{\boldsymbol{\xi}}\  T(\boldsymbol{\xi})+J_{\mathrm{coll}}(\boldsymbol{\xi}) + J_{\mathrm{joint}}(\boldsymbol{\xi})\Comma
\end{align}
where $J_{\mathrm{coll}}$ is a penalty to avoid collisions and $J_{\mathrm{joint}}$ aims to avoid joint limit violations. Therefore, the trajectories are time-optimal in the sense that the algorithm aims to find the shortest possible trajectory that can be represented by $N_{\mathrm{via}}$ via-points.

\subsubsection{Time parameterization with TPFRC}\label{TOPP-TPFRC}
Time parameterization involves finding \(s(t)\) for a given path \(\hat{\mathbf{q}}_A(s)\), with \(s \in [0,1]\), such that \(\mathbf{q}_A(t) = \hat{\mathbf{q}}_A\big(s(t)\big)\). Existing methods generally assume that \(\dot{s}(t) \geq 0\) always and \(\dot{s}(t) > 0\) almost everywhere \cite{pham:2014, verscheure:2009, pham:2018}, where the linear time parametrization of the VP-STO can be considered a special case. To incorporate the PFRC into VP-STO, we demonstrate that it shares a similar structure to the velocity constraints expressed in (\ref{eq:TPVelAcc}). This structural similarity allows the PFRC to be integrated into various parameterization techniques \cite{pham:2014, verscheure:2009, pham:2018}, as well. Using the differential kinematics from \eqref{eq:HydraulicVelocities} with \eqref{eq:TPVelAcc}, the cylinder speeds are given by
\begin{equation}\label{eq:ddot}
\begin{aligned}
    \dot{\mathbf{d}}&=\mathbf{J}_C(\hat{\mathbf{q}}_A)\hat{\mathbf{q}}_A^{\prime}\dot{s}=\hat{\mathbf{d}}^\prime(s)\dot{s}\ .
\end{aligned}
\end{equation}
Since \(\dot{s} > 0\) almost everywhere, we can deduce from \eqref{eq:ddot} that \(\mathrm{sign}(\dot{\mathbf{d}}) = \mathrm{sign}(\hat{\mathbf{d}}^\prime)\) almost everywhere. If \(\dot{s} = 0\), then \eqref{eq:ddot} implies that \(\dot{\mathbf{d}} = \mathbf{0}\), and from \eqref{eq:TotalPumpFlow}, it follows that \(Q = 0\). Consequently, any case where \(\mathrm{sign}(\dot{\mathbf{d}}) \neq \mathrm{sign}(\hat{\mathbf{d}}^\prime)\) does not impact the result. Thus, for all \(\dot{s} \geq 0\), we have
\begin{align}\label{eq:TOPPRRTSumFlowRate}
    Q&= \hat{Q}(s)\dot{s}\Comma
\end{align}
where \(\hat{Q}(s) = \sum_{l=1}^{n_A} A_l \big( \mathrm{sign}(\hat{d}_{l}^{\prime}) \big) |\hat{d}_l^\prime(s)|\). This proves that the PFRC is consistent with the velocity constraints, allowing for its seamless integration into the VP-STO framework.

\subsubsection{Collision cost}
For collision avoidance the original work \cite{janakowski:2023} proposes binary costs $J_{\mathrm{coll}}(\boldsymbol{\xi})=J_{\mathrm{bin}}(\boldsymbol{\xi})$  with
\begin{align}\label{eq:VPSTOCollisionCostBinary}
    J_{\mathrm{bin}}(\boldsymbol{\xi})&= w_{\mathrm{coll}}\sum_{k=0}^{N_{\mathrm{eval}}}\sum_{(i,j)\in\mathcal{I}_c}\mathbb{I}\Big(\mathrm{sd}\big(\mathcal{L}_i(\hat{\mathbf{q}}_{A,k}),\mathcal{O}_j\big)<0\Big)\Comma
\end{align}
where $\mathbb{I}$ is the indicator function. As stated in \cite{janakowski:2023}, this has the advantage that binary collision checking is computationally cheaper than distance computation. However, we propose a different cost formulation, which we refer to the \textit{weighted signed distance function} (W-SDF) cost, i.e. we use $J_{\mathrm{coll}}(\boldsymbol{\xi})=J_{\mathrm{WSDF}}(\boldsymbol{\xi})$
\begin{align}\label{eq:VPSTOCollisionCostWSDF}
    J_{\mathrm{WSDF}}(\boldsymbol{\xi})&=\sum_{k=0}^{N_{\mathrm{eval}}}\sum_{(i,j)\in\mathcal{I}_c}w_{\mathrm{coll},i}l_{\mathrm{coll},ij}(\hat{\mathbf{q}}_{A,k})\Comma
\end{align}
with
\begin{align}
    l_{\mathrm{coll},ij}(\hat{\mathbf{q}}_{A,k})=\begin{cases}
        1-\mathrm{sd}_{ij}(\hat{\mathbf{q}}_{A,k}), & \text{if } \mathrm{sd}_{ij}(\hat{\mathbf{q}}_{A,k}) \leq 0\\
        0, &\text{otherwise.}
    \end{cases}
\end{align}
Here, $w_{\mathrm{coll},i}$ is a link-dependent weight that decreases with the link number $i$. For constant $w_{\mathrm{coll},i}$, this corresponds to a penalty of the penetration depth, which we refer to as the signed distance function (SDF) cost.

The reasoning for the link-dependent weights is as follows: Links earlier in the kinematic chain have fewer degrees of freedom (DoF) available for avoiding obstacles compared to their child links. To achieve a collision-free configuration, adjustments must be made to the joints that affect a given link so that both, the link itself and its child links remain free of collisions. In contrast, the joints of child links can be adjusted without affecting the collision state of the parent link. Therefore, link-dependent weights, $w_{\mathrm{coll},i}$, are used to reflect the fact that it is more challenging to resolve a collision involving an earlier link in the chain than one involving a child link. Additionally, a signed distance function is employed to indicate that a minor contact with an obstacle can be more easily resolved to a collision-free state than a deeper penetration. We argue that this added distinction between "better" and "worse" collisions helps the CMA-ES algorithm compute more effective updates toward collision-free trajectories. In order to satisfy the joint limits, we use the same penalty costs as proposed in the original work \cite{janakowski:2023}.

\subsection{Local Planner}
The local planner aims to track the reference of the global planner while considering the motions of the passive joints due to (\ref{eq:ddqu_DynamicsSimplified}). We formulate a discrete-time optimal control problem (OCP) of the form
\begin{subequations}\label{eq:DiscreteTimeOCPProblem}
\begin{align}
    \min\limits_{\mathbf{X},\mathbf{U}} &\sum_{k=i}^{i+N-1}l_k(\mathbf{x}_k,\mathbf{u}_k)\label{eq:DiscreteTimeOCPProblem objective}    \\
    \text{s.t.}\;  &\mathbf{x}_{k+1}=\mathbf{f}(\mathbf{x}_k,\mathbf{u}_k) \ ,\; \mathbf{x}_i=\mathbf{x}_{\mathrm{init},i}
    \label{eq:OCPDynamics}\\
    & \mathbf{q}_k\in[\mathbf{q}_{\mathrm{max}},\mathbf{q}_{\mathrm{min}}]\ ,\; Q\leq Q_{\mathrm{max}}\label{eq:OCPlimitsOCPvolumeFlowConstaints}
    \\
    & \mathbf{u}_k\in[\mathbf{u}_{\mathrm{min}},\mathbf{u}_{\mathrm{max}}]\\
    &    \mathrm{sd}_{ij}(\mathbf{q})> 0 \ , (i,j)\in \mathcal{I}_c    \label{eq:OCPInequalities}\ ,
\end{align}
\end{subequations}
with $\mathbf{X}=[\mathbf{x}_i,\dots,\mathbf{x}_{i+N-1}]$, $\mathbf{U}=[\mathbf{u}_i,\dots,\mathbf{u}_{i+N-1}]$, the state vector $\mathbf{x}_k^{\mathrm{T}}=[\mathbf{q}_k^{\mathrm{T}},\dot{\mathbf{q}}_k^{\mathrm{T}}]\in\mathds{R}^{2n}$, the control input $\mathbf{u}_k=\ddot{\mathbf{q}}_{A,k}\in\mathds{R}^{n_A}$, where the notation $\mathbf{x}_k=\mathbf{x}(k T_s)$ with the sampling time $T_s$ is used. The local planner is executed on a moving horizon where the initial condition $\mathbf{x}_{\mathrm{init},i}$ is obtained by integration of (\ref{eq:ddqu_DynamicsSimplified}) along the solution $\mathbf{u}_{i-1}^*$ of the previous iteration with initial condition $\mathbf{x}_{\mathrm{init},i-1}$. Contrary to \cite{jebellat:2023}, we include collision, joint, and acceleration limit constraints as well as the PFRC in the OCP of the local planner.

The intermediate cost $l_k$ in (\ref{eq:DiscreteTimeOCPProblem objective}) comprises four parts, i.e.,
\begin{align}
\begin{aligned}
    l_k(\mathbf{x}_k,&\mathbf{u}_k)=\|\mathbf{q}_{A,k}-\mathbf{q}_{A,k}^{\mathrm{ref}}\|_2^2
    +w_1\|\mathbf{u}_k-\ddot{\mathbf{q}}_{A,k}^{\mathrm{ref}}\|_2^2\\
    &+w_2\|\mathbf{q}_{P,k}-{\mathbf{q}}_{P,k}^{\mathrm{ref}}\|_2^2+w_3\|\dot{\mathbf{q}}_{P,k}-\dot{\mathbf{q}}_{P,k}^{\mathrm{ref}}\|_2^2\ ,
\end{aligned}
\end{align}
with weighting parameters $w_1,w_2,w_3>0$. The first and second parts penalize deviations from the trajectory ${\mathbf{q}}_{A,k}^{\mathrm{ref}}$ and the corresponding accelerations $\ddot{\mathbf{q}}_{A,k}^{\mathrm{ref}}$ of the actuated joints derived by the global planner VP-STO ensures that the solver aims to find a solution close to the VP-STO solution For the passive joints, the equilibrium point trajectory ${\mathbf{q}}_{P,k}^{\mathrm{ref}}=\mathbf{g}_P^{-1}({\mathbf{q}}_{A,k}^{\mathrm{ref}})$ is computed in advance. To dampen the motions of the passive joints, the third and fourth parts are included, which penalize deviations from the equilibrium points and their change over time $\dot{\mathbf{q}}_{P,k}^{\mathrm{ref}}$. The OCP \eqref{eq:DiscreteTimeOCPProblem} is solved using an augmented Lagrangian constrained iLQR algorithm \cite{howell:2019}. 



\begin{table}
    \centering
    \caption{Runtimes of the PFRC-VP-STO planner in 100 runs.}\label{tab:RuntimesVPSTO}
    \begin{tabular}{||c|c||c|c|c||}
        \hline
         \multicolumn{1}{||c|}{\textbf{Env.}} & \textbf{Log} &  \multicolumn{3}{|c||}{\textbf{Runtime [s]}}\\
         \hline\hline
         &&Binary & SDF & W-SDF\\
        \hline
        1 & 1 (p)
                &0.90$\pm$0.09  & 0.99$\pm$0.11 & 0.98$\pm$0.12\\
        1 & 1 (l)
                & 1.78$\pm$0.27 & 2.00$\pm$0.30 & 2.10$\pm$0.83\\
        \hline
        1 & 2 (p) 
                  & 1.22$\pm$0.20 & 1.36$\pm$0.25 & 1.35$\pm$0.26\\
        1 & 2 (l) 
                  & 1.96$\pm$0.50 & 2.23$\pm$0.73 &2.17$\pm$0.55\\
        \hline
        1 & 3 (p) 
                  & - & 4.14$\pm$2.00 &3.25$\pm$0.60\\
        1 & 3 (l) 
                  & - & - &4.95$\pm$1.44\\
        \hline\hline
        2 & 1 (p) 
                  & 0.94$\pm$0.10 & 1.00$\pm$0.08 & 1.01$\pm$0.1\\
        2 & 1 (l) 
                  & 2.34$\pm$0.89 & 2.81$\pm$1.61 & 2.93$\pm$1.03\\
        \hline
        2 & 2 (p) 
                      & - & - & 2.64$\pm$1.15\\
        2 & 2 (l) 
                  & - & - & 6.38$\pm$1.28\\
        \hline
        2 & 3 (p) 
                  & 1.84$\pm$0.53 & 2.02$\pm$0.68 & 2.15$\pm$0.71\\
        2 & 3 (l) 
                  & 3.71$\pm$1.64 & 3.85$\pm$1.30 & 3.82$\pm$1.04\\
        \hline
    \end{tabular}
\end{table}
\begin{figure}[htb!]
\centering
\adjustbox{trim=3.9cm 0.8cm 3cm 1.9cm, clip}{\includegraphics[scale=0.28]{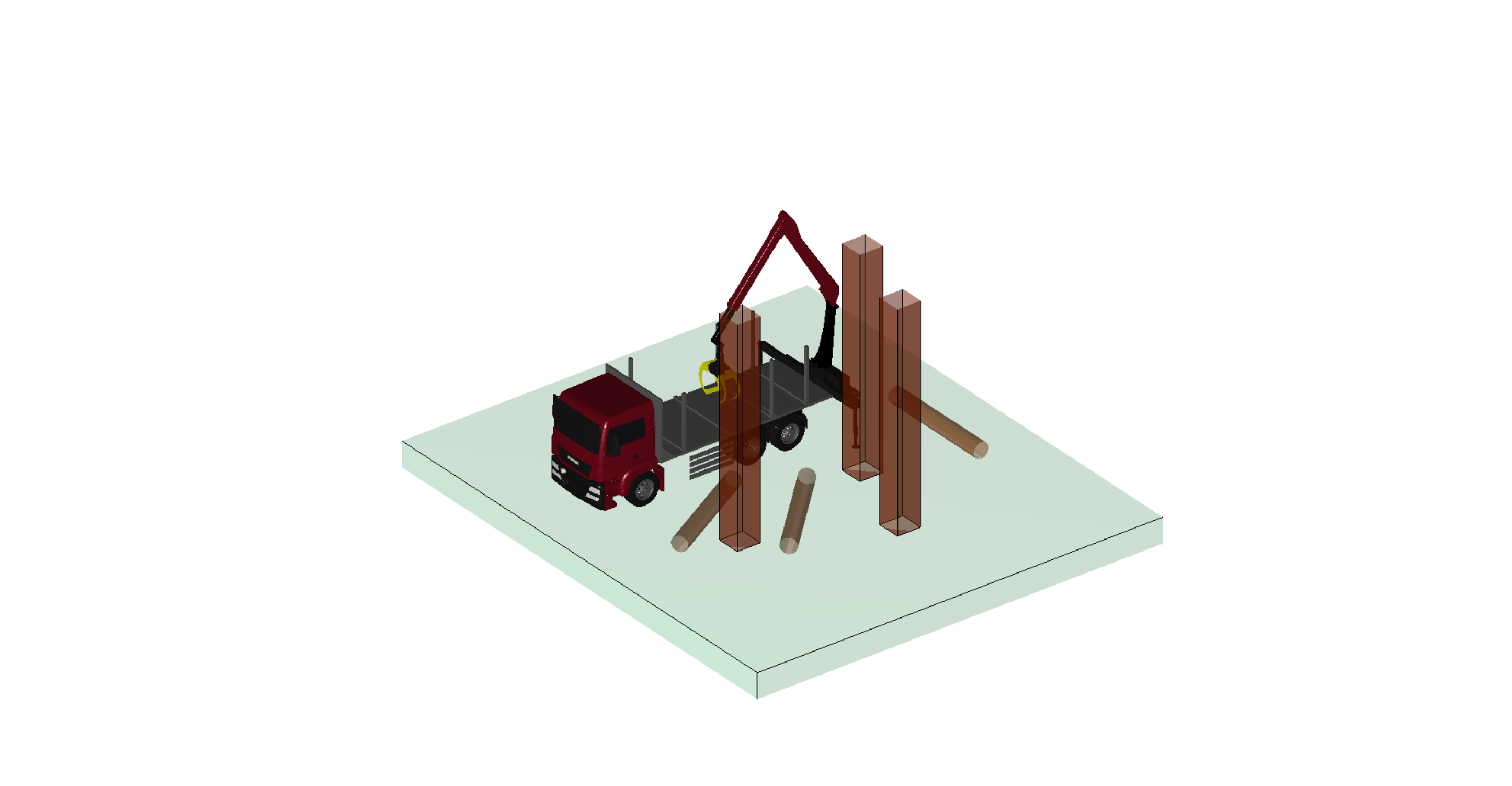}}
\adjustbox{trim=3.9cm 0.8cm 3cm 1.9cm, clip}{\includegraphics[scale=0.28]{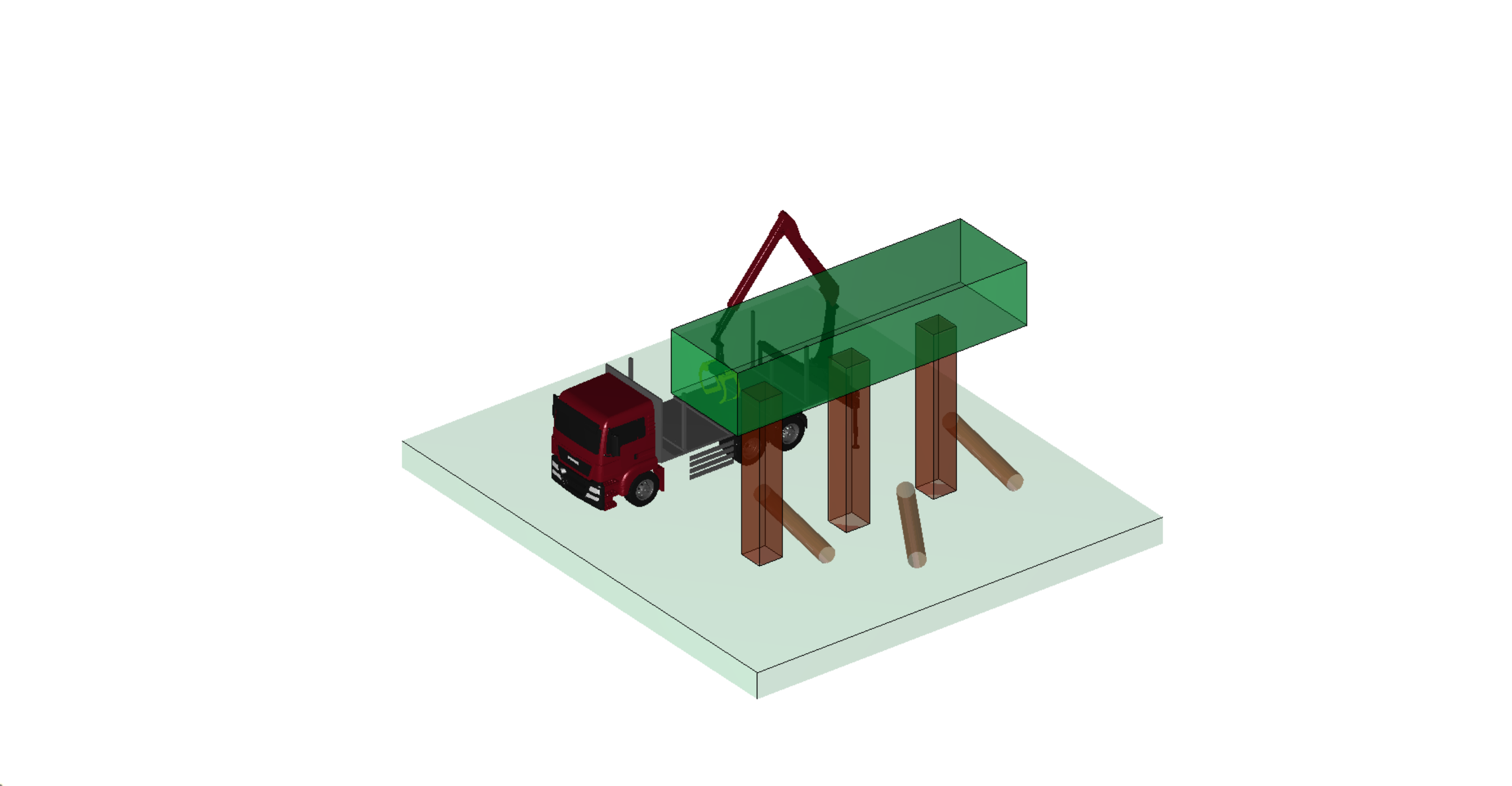}}
\caption{Environments used in the experiments: Environment 1 (top) and Environment 2 (bottom) with logs 1-3 (left to right)}
\label{fig:Environments}
\end{figure}
\begin{table*}
    \centering
    \caption{Success rates and trajectory durations of the global planners in 100 runs.}\label{tab:StatisticsVPSTO}
    \begin{tabular}{||c|c||c|c|c|c||c|c|c|c||}
        \hline
         \multicolumn{1}{||c|}{\textbf{Env.}} & \textbf{Log} & \multicolumn{4}{|c||}{\textbf{Success Rate [\%]}} & \multicolumn{4}{|c||}{\textbf{Traj. Duration [s]}}\\
         \hline\hline
         &&RRT* & Binary & SDF & W-SDF & RRT* & Binary & SDF & W-SDF\\
        \hline
        1 & 1 (p)& 94 & 100 & 100 & 100
                & 9.50$\pm$0.91 & 6.29$\pm$0.02 & 6.30$\pm$0.03 & 6.29$\pm$0.02\\
        1 & 1 (l) & 91 & 100 & 100 & 100
                  & 8.37$\pm$0.68 & 6.67$\pm$0.03 & 6.67$\pm$0.04 & 6.67$\pm$0.03\\
        \hline
        1 & 2 (p) & 100 & 100 & 100 & 100
                  & 8.50$\pm$0.97 & 5.42$\pm$0.09 & 5.41$\pm$0.08 & 5.42$\pm$0.07\\
        1 & 2 (l) & 92 & 100 & 100 & 100
                  & 8.20$\pm$0.74 & 5.64 $\pm$0.04 & 5.64$\pm$0.04 & 5.64$\pm$0.03\\
        \hline
        1 & 3 (p) & 100 & 0 & 11 & 100
                  & 19.24$\pm$1.82 & - & 17.29$\pm$0.23 & 17.49$\pm$ 0.49\\
        1 & 3 (l) & 0 & 0 & 0 & 87
                  & - & - & - & 18.43$\pm$0.35\\
        \hline\hline
        2 & 1 (p) & 99 & 100 & 100 & 100
                  & 8.61$\pm$0.58 & 6.74$\pm$0.01 & 6.74$\pm$0.01 & 6.74$\pm$0.01\\
        2 & 1 (l) & 73 & 100 & 100 & 100
                  & 9.98$\pm$1.27 & 7.10$\pm$0.07 & 7.09$\pm$ 0.05 & 7.06$\pm$0.08\\
        \hline
        2 & 2 (p) & 13 & 0 & 0 & 92 
                  & 8.47$\pm$ 1.13 & - & - & 10.34$\pm$0.80\\
        2 & 2 (l) & 57 & 0 & 0 & 79
                  & 7.99$\pm$0.73 & - & - & 10.49$\pm$0.48\\
        \hline
        2 & 3 (p) & 100 & 99 & 100 & 100
                  & 11.86$\pm$1.35 & 7.50$\pm$1.06 & 7.33$\pm$0.12 & 7.66$\pm$1.12\\
        2 & 3 (l) & 75 & 100 & 100 & 100
                  & 14.21$\pm$1.97& 8.30$\pm$0.21 & 8.29$\pm$0.31 & 8.91$\pm$1.79\\
        \hline
    \end{tabular}
\end{table*}

\begin{figure*}[ht!]
\centering
\adjustbox{trim=1.4cm 0.2cm 1.4cm 0.3cm,clip}{\includegraphics[scale=0.65]{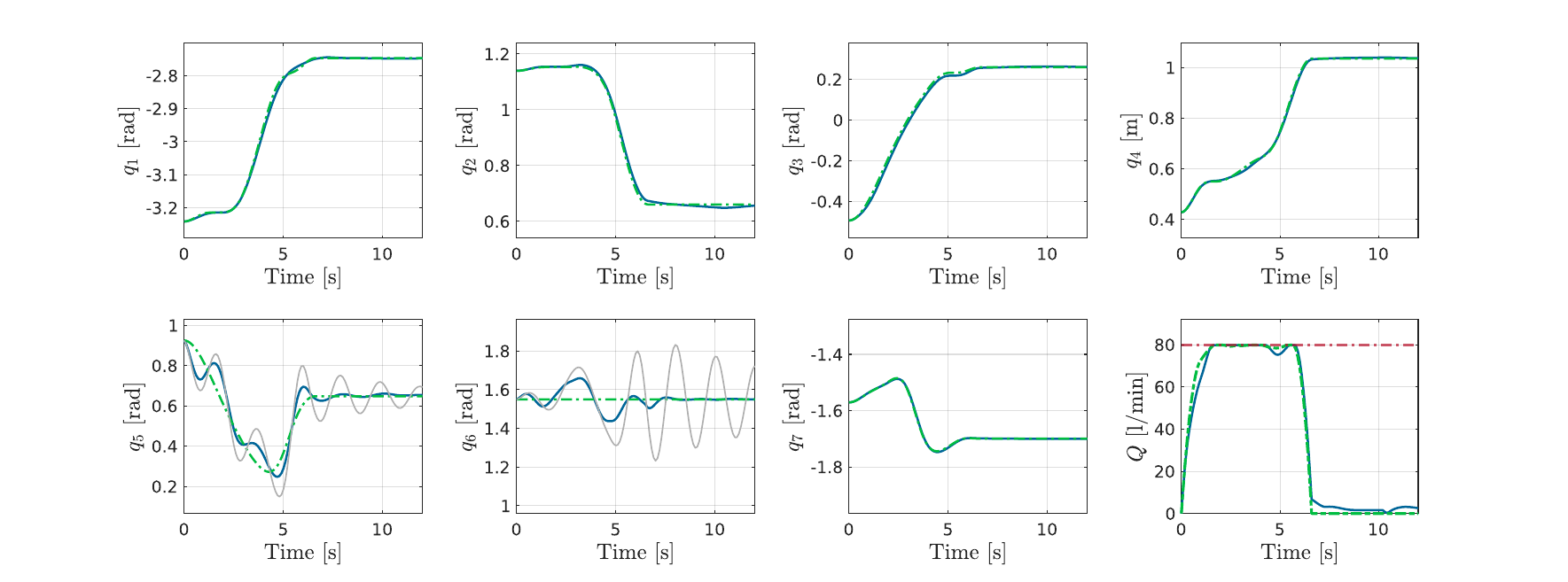}}
\caption{Numerical results for a sample trajectory: Proposed hybrid motion planner with PFRC VP-STO based global planner + iLQR based local planner (blue lines), reference of PFRC-VP-STO (green dashed line), PFRC VP-STO applied to the system with passive dynamics (\ref{eq:ddqu_DynamicsSimplified}) without local planner (gray line), pump flow rate limit (red dashed line).  }
\label{fig:PumpFlowFinal}
\end{figure*}

\section{Numerical Results}\label{sec:Results}
We evaluate the algorithms in two different environments, as shown in Figure~\ref{fig:Environments}. In each environment, the crane must perform point-to-point motions to load three logs onto the truck. For each log, the crane must first move to pick it up (pick p) and then transport it to the truck (loading l), resulting in a total of 12 scenarios. It is important to note that during the loading process, the log itself becomes an additional collision obstacle, and the mass parameters of the passive dynamics change when the crane is carrying a log. The global planning capabilities of the PFRC-VP-STO as an optimal single query planner are evaluated by comparing the performance to an informed RRT* algorithm \cite{gammel:2014}, a state-of-the-art optimal single query planner. To obtain a TOPP for the RRT* path, we use the approach described in \cite{verscheure:2009} together with the PFRC from Section~\ref{TOPP-TPFRC}.
All algorithms are implemented in \textsc{C++} and all experiments are performed on an Intel Core i7-10850H CPU @ 2.7 GHz $\times$ 12. We use the \textit{Flexible Collision Library (FCL)} \cite{pan:2012} for collision detection and distance computation. The gradients of the collision constraints in the local planner are computed as described in \cite{schulman:2014}. As in \cite{ayoub:2024}, we use the \textit{Open Motion Planning Library (OMPL)} \cite{sucan:2012} implementation of the informed RRT*.


\subsection{Global Planner}
Due to the stochasticity of the planners, we perform 100 runs for every scenario.
Table~\ref{tab:StatisticsVPSTO} presents the success rates and mean trajectory durations (± standard deviation) for the TOPP-RRT* and PFRC-VP-STO algorithms under different collision cost formulations: "Binary" (\ref{eq:VPSTOCollisionCostBinary}), "SDF" (\ref{eq:VPSTOCollisionCostWSDF}) with a constant weight \( w_{\mathrm{coll},i} = 10^2 \), and "W-SDF" (\ref{eq:VPSTOCollisionCostWSDF}) with \( w_{\mathrm{coll},2} = 10^5 \), \( w_{\mathrm{coll},3} = 10^4 \), \( w_{\mathrm{coll},4} = 10^3 \), and \( w_{\mathrm{coll},5} = 10^2 \). We determined that using \( N_{\mathrm{via}} = 6 \) via-points with a CMA-ES population size of $50$ and an initial sample covariance of $\sigma_{\mathrm{init}}=1$ provides a good balance between runtime and time optimality. The mean runtimes and standard deviations for PFRC-VP-STO are provided in Table~\ref{tab:RuntimesVPSTO}. The RRT* algorithm is stopped after 10 seconds, which is longer than any successful run of the PFRC-VP-STO.

With binary collision costs, PFRC-VP-STO fails to find a solution for reaching and loading log 3 in environment 1, as well as for log 2 in environment 2, because the CMA-ES algorithm lacks sufficient information to perform effective updates. Using the SDF formulation slightly improves the success rates for reaching log 3 in environment 1. The proposed W-SDF method tremendously increase the success rates compared to the Binary and SDF costs.

As shown by the success rates in Table~\ref{tab:StatisticsVPSTO}, the RRT* planner is less robust than PFRC-VP-STO with W-SDF costs within the 10-second time limit. Notably, for the task of loading log 3, the informed RRT* planner fails to find any solution within 10 seconds over 100 runs, highlighting the difficulty of this task. Furthermore, the trajectory durations in Table~\ref{tab:StatisticsVPSTO} indicate that PFRC-VP-STO consistently finds faster trajectories in almost all cases. In addition, the standard deviation of the trajectory duration of the PFRC-VP-STO is lower than those computed by the TOPP of the RRT* path. The green dashed lines in Figure~\ref{fig:PumpFlowFinal} illustrate a sample trajectory with the PFR, demonstrating that the proposed global planning algorithm computes reference trajectories that operate near the limit of the available PFR.

\subsection{Local Planner}
We evaluate our hybrid motion planning strategy on 100 global trajectories for each scenario. The local planner operates at an update rate of 10 Hz (i.e., \(T_s = 0.1\) s) with a horizon length of \(N = 40\) steps (i.e., 4 seconds). Through experimentation, we determined that weights \(w_1 = 0.5\), \(w_2 = 0.005\), and \(w_3 = 0.05\) provide a good balance between reference tracking and sway damping. The local planner succeeded in all experiments, reaching and stopping at the desired target configuration while satisfying all constraints. The runtime for each iteration remained below \(T_s = 0.1\) s, which is suitable for a real-time implementation.

Figure~\ref{fig:PumpFlowFinal} illustrates a sample trajectory for log 1 in environment 1. The reference trajectory of the proposed global planner is depicted in green dashed lines, the result of the proposed hybrid planner (PFRC VP-STO + iLQR based local planner)  in blue lines, and the motions of the passive joints, obtained by directly integrating (\ref{eq:ddqu_DynamicsSimplified}) along the reference trajectory of the global planner, are shown in gray. It is clearly visible that the local planner effectively follows the reference, dampens the motions of the passive joints, and satisfies all constraints, including collision constraints involving the gripper due to the passive dynamics. Furthermore, the corresponding PFR shown in Figure~\ref{fig:PumpFlowFinal}, indicates that the final trajectory of the proposed hybrid motion planning strategy remains near the limit, which indicates that the planned trajectory is close to time-optimality. When this constraint is not enforced, the PFRC would be violated. Moreover, when the PFRC is considered without collision constraints, collisions occur, particularly with the truck walls. This shows the relevance of considering both constraints for the motion planning of the hydraulically actuated timber crane.

\section{Conclusion \& Future Work}\label{sec:conclusion}
This work presents a near time-optimal hybrid motion planning framework for a hydraulically driven timber crane with two passive joints. We extended the VP-STO algorithm to incorporate pump flow rate constraints and demonstrated its near time-optimal global planning capabilities. Additionally, we showed that a gradient-based local planner can effectively track the reference from the global planner while accounting for the motions of the passive joints, ensuring both collision avoidance and sway damping of the nominal system. Current work focuses on investigating the approach under imperfect tracking conditions of the subordinate controller, with the goal of deploying it on the real machine.

\bibliographystyle{plain} 
\bibliography{refs} 

\end{document}